\documentclass[runningheads]{llncs}
\usepackage{graphicx}
\usepackage{hyperref}
\hypersetup{colorlinks=true,
    linkcolor=green,
    filecolor=magenta,      
    urlcolor=blue,
}
\usepackage{url}
\usepackage{color}
\newcommand{\ttt}[1]{\texttt{#1}}
\newcommand{\knn}{$k$-NN}
\newcommand{\modelone}{Model~1}
\newcommand{\flexneuart}{\texttt{FlexNeuART}}

\begin{document}
\title{Traditional IR rivals neural models on the MS~MARCO Document Ranking Leaderboard}
%
%
\author{Leonid Boytsov \\\email{leo@boytsov.info}
}
\institute{}
\maketitle              
\begin{abstract}
This short document describes a traditional IR 
system that achieved MRR@100 equal to 0.298
on the MS MARCO Document Ranking leaderboard (on 2020-12-06).
Although inferior to most BERT-based models, 
it outperformed several neural runs (as well as all non-neural ones), 
including two submissions that used a large pretrained Transformer model for re-ranking.
We provide software and data to reproduce our results.
\end{abstract}
\section{Introduction and Motivation}
A typical text retrieval system uses a multi-stage retrieval pipeline,
where documents flow through a series of ``funnels`` 
that discard unpromising candidates
using increasingly more complex and accurate ranking components.
These systems have been traditionally relying on simple term-matching techniques
to generate an initial list of candidates \cite{buttcher2016information,croft2010search}.
In that, retrieval performance is adversely affected by a mismatch between query and document terms,
which is known as a vocabulary gap problem~\cite{furnas1987vocabulary,zhao2010term}.

The vocabulary gap can be mitigated by learning  dense or sparse representations for effective
first-stage retrieval.
Despite recent success in achieving this objective \cite{lee2019latent,karpukhin2020dense,xiong2020approximate}, existing studies have have at least one of the following flaws: 
\begin{itemize}
    \item They often compare against a weak baseline such as untuned BM25 \cite{Robertson2004};
    \item They nearly always rely on \emph{exact} brute-force \knn\   search,
      which is not exactly practical.
      In that, they mostly ignore efficiency-effectiveness and scalability
      trade-offs related to using \emph{approximate} \knn\   search (\S 3.3 \mbox{in~\cite{boytsov2018efficient}}).
    \item They require expensive index-time computation using a large Transformer~\cite{vaswani2017attention} model.
\end{itemize}

This motivated us to develop a carefully-tuned traditional, i.e., non-neural, system, which we evaluated in
\href{https://microsoft.github.io/MSMARCO-Document-Ranking-Submissions/leaderboard/}{the MS MARCO document ranking task} \cite{nguyen2016ms,craswell2020overview}. 
Our objectives are:
\begin{itemize}
    \item To provide a stronger traditional baseline;
    \item To develop an effective first-stage retrieval system, 
          which can be efficient and effective without expensive index-time precomputation.
\end{itemize}

Our submission (dated 2020-12-06) achieved MRR=0.298 on the hidden validation set 
and outperformed all other traditional systems.
It was the first system (on this leaderboard) that outstripped several neural baselines. 
According to our own evaluation 
on TREC NIST data \cite{craswell2020overview},
our system achieves NDCG@10 equal
to 0.584 and 0.558 on  2019 and 2020 queries, respectively.
It, thus, outperforms a tuned BM25 system by 6-7\%:
NDCG@10 is equal to 0.544 and 0.524 on 2019 and 2020 queries, respectively.
\href{https://github.com/oaqa/FlexNeuART/blob/repr2020-12-06/scripts/data_convert/msmarco/README.md}{We posted two notebooks to reproduce results}:
\begin{itemize}
    \item \href{https://github.com/oaqa/FlexNeuART/blob/repr2020-12-06/scripts/data_convert/msmarco/MSMARCO_docs_2020-12-06_complete.ipynb}{The first notebook} reproduces all steps necessary to download the data, preprocess it, and train all the models.
    \item \href{https://github.com/oaqa/FlexNeuART/blob/repr2020-12-06/scripts/data_convert/msmarco/MSMARCO_docs_2020-12-06_processed_data_and_precomp_model1.ipynb}{The second notebook} operates on preprocessed data in \flexneuart\   JSONL format. It does not require running MGIZA to generate IBM Model 1 (these models are already trained).
\end{itemize}

\section{System Description}
\subsection{Overview}
We use our \flexneuart\  retrieval
toolkit\cite{boytsov2020flexible}\footnote{\url{https://github.com/oaqa/FlexNeuART}}, 
which ingests data (queries and documents) in the form of \emph{multi-field} JSON entries.
A field can include any textual information associated with a document.
In particular, each MS~MARCO document has three text attributes: 
\ttt{URL}, \ttt{title}, and \ttt{body} (main document text with HTML formatting removed). 
A field can be provided as is, i.e., without any modification, 
or it can be preprocessed and split into tokens.
Given an original textual attribute, e.g., the \ttt{title}, it is possible to generate multiple fields,
which are tokenized, lemmatized, stopped, or otherwise processed in different ways. 

In the current system,
a textual attribute of an MS MARCO document
can be tokenized using Spacy\cite{spacy2} or split into BERT word pieces \cite{devlin2018bert,WuSCLNMKCGMKSJL16}.
The tokenized field can be further lemmatized and/or stopped (stopping is not applied to BERT word pieces).
In the case of \ttt{URL}, we heuristically preprocess input before applying Spacy:
We remove prefixes such as \texttt{http://}, slashes, and other punctuation signs.

Our submission retrieves 1000 candidate documents using Apache Lucene\footnote{\url{https://lucene.apache.org/}} 
with a tuned BM25 \cite{Robertson2004} and re-ranks documents 
using an LambdaMart \cite{burges2010ranknet} model, 
which aggregates 13 features described below.

\subsection{Features}

Overall we compute 13 features,
which include simple textual similarity features as well as the lexical translation features,
namely, IBM \modelone\ log-scores \cite{berger1999information}.
\href{https://github.com/oaqa/FlexNeuART/blob/repr2020-12-06/scripts/data_convert/msmarco/exper_desc.lb2020-12-04/extractors/best_classic_ir_expand_full.json}{The corresponding JSON descriptor can be found online}.
BM25 and IBM \modelone \cite{berger1999information} log-scores are core ingredients.
The full set of features includes:
\begin{itemize}
    \item BM25 \cite{Robertson2004} scores, which are \emph{normalized} using the sum of query term IDFs;
    \item cosine similarity scores;
    \item relative query-document overlap scores: the number of query terms that appear in documents,
    \emph{normalized} by the total number of query terms.
    \item \emph{normalized} proximity scores (similar to ones used in our prior TREC experiments \cite{boytsov2011evaluating}): Each scoring component generates two feature values;
    \item IBM \modelone \cite{berger1999information} \emph{log}-scores, 
        which are \emph{normalized} by the number of query terms.
\end{itemize}

IBM \modelone \cite{berger1999information} is a lexical translation model, 
which is trained using a \emph{parallel corpus} (a bitext).
The parallel corpus consists of queries paired with respective relevant documents.
Training relies on the expectation-maximization algorithm \cite{brown1993mathematics} implemented in MGIZA \cite{Och2003}.\footnote{\url{https://github.com/moses-smt/mgiza/}}

Overall we compute four IBM \modelone\ log-scores. 
Three scores are computed for \emph{non}-lemmatized attributes: \ttt{URL}, \ttt{title}, and \ttt{body},
which are tokenized with Spacy\cite{spacy2}.
One score is computed for \ttt{body}, which is split into BERT word pieces~\cite{devlin2018bert,WuSCLNMKCGMKSJL16}.

We take several measures to maximize the effectiveness of IBM \modelone\ (see
\S~3.1.1.2 \cite{boytsov2018efficient} for details).
Most importantly, it is necessary to have queries and documents of comparable sizes.
It is true for \ttt{URL} and \ttt{title}, but not for \ttt{body}, which typically has hundreds of tokens.
To resolve this issue, for each pair of query $q$ and respective relevant document $d$, 
we first split $d$ into multiple short chunks $d_1$, $d_2$, \ldots $d_n$.
Then, we replace the pair $(q, d)$ with a set of pairs $\{ (q, d_i) \}$.

\subsection{Efficiency Considerations}
Although our traditional system is reasonably efficient, it is research software,
which lags somewhat in speed compared to an optimized traditional system.
In particular, it contains an inefficient module computing proximity score in time $O(|D| \times w)$,
where $w$ is the size of the window (to count word pairs) and $|D|$ is the document length.
It can be implemented to run $w$ times faster.

Furthermore, we can improve computation of IBM \modelone.
First, translation tables produced by MGIZA can be pruned more aggressively with a small loss in accuracy.
Second, the algorithm runs in time $O(|D|\times|q|)$, where $|q|$ and $|D|$ are query and document lengths.
This can be reduced to $O(|D|)$ by precomputing a query-specific reverse-translation table 
(\S~3.1.2.1 \cite{boytsov2018efficient}). 
Precomputation requires extra time, 
but it is totally worth the effort when we need to re-rank a large number of documents. 

Last but not least, re-ranking requires \emph{random}-access to data stored in a forward index.
Reading forward-index entries from memory can be quite fast.
However, storing several fields per documents can be space inefficient.
At the same time, if data is read from a disk, reading can become a primary computational bottleneck.
\flexneuart\  generates and reads forward indices for each field separately,
because it is a simpler and more flexible approach.
In a production system, all field entries belonging to the same document can be stored jointly. 
Because recent consumer SSD drives can sustain about 25K random reads 
per second\footnote{\url{https://ssd.userbenchmark.com/SpeedTest/693540/Samsung-SSD-970-EVO-Plus-1TB}},
it is quite practical to re-rank hundreds of documents per query
by reading forward-index entries from a fast SSD disk rather than from main memory.


\end{document}